\documentclass[10pt,twocolumn,letterpaper]{article}

\usepackage{iccv}
\usepackage{times}
\usepackage{epsfig}
\usepackage{graphicx}
\usepackage{amsmath}
\usepackage{amssymb}
\usepackage[accsupp]{axessibility}

\usepackage[pagebackref=true,breaklinks=true,letterpaper=true,colorlinks,bookmarks=false]{hyperref}

\iccvfinalcopy 

\def\iccvPaperID{9976} 

\ificcvfinal\fi
\graphicspath{{figures/}}

\newcommand{\Eq}[1]{Eq.~(\ref{eq:#1})}
\newcommand{\eq}[1]{\Eq{#1}}
\newcommand{\fig}[1]{Fig.~\ref{fig:#1}}
\newcommand{\tab}[1]{Tab.~\ref{tab:#1}}

\newcommand{\sect}[1]{Section~\ref{sec:#1}}

\def\argmin{\operatornamewithlimits{\rm arg\,min}}

\usepackage{arydshln}

\usepackage{xcolor}
\usepackage{graphicx}
\usepackage{amsmath}
\usepackage{mathtools}  
\usepackage{amsfonts} 
\usepackage{amssymb}
\usepackage{booktabs}
\usepackage{comment}
\usepackage{dblfloatfix} 

\usepackage{xspace}

\begin{document}

\title{A step towards understanding why classification helps regression}

\author{Silvia L. Pintea$^{1,2}$
\and
Yancong Lin$^{3}$
\and
Jouke Dijkstra$^{2}$
\and
Jan C. van Gemert$^{1}$
\and
{\normalsize $^{1}$ Computer Vision Lab, Delft University of Technology}\\
{\normalsize $^{2}$ Division of Image Processing (LKEB), Leiden University Medical Center}\\
{\normalsize $^{3}$ Intelligent Vehicles Group, Delft University of Technology
}\\
}
\maketitle
\ificcvfinal\thispagestyle{empty}\fi
\begin{abstract}
A number of computer vision deep regression approaches report improved results when adding a classification loss to the regression loss.
Here, we explore why this is useful in practice and when it is beneficial. 
To do so, we start from precisely controlled dataset variations and data samplings and find that the effect of adding a classification loss is the most pronounced for regression with imbalanced data.
We explain these empirical findings by formalizing the relation between the balanced and imbalanced regression losses.
Finally, we show that our findings hold on two real imbalanced image datasets for depth estimation (NYUD2-DIR), and age estimation (IMDB-WIKI-DIR), and on the problem of imbalanced video progress prediction (Breakfast).
Our main takeaway is: for a regression task, if the data sampling is imbalanced, then add a classification loss.
\end{abstract}

\section{Introduction}

Regression models predict continuous outputs.
In contrast, classification models make discrete, binned, predictions. 
For a continuous task, regression targets are a super-set of the classification labels: they are more precise, taking values in-between the discrete classification bins. 
For regression, the error is only bounded by the precision of the measurements, for classification this also depends on the bin sizes: e.g. an age estimation classifier that can predict only young/old classes, cannot discriminate between middle-aged people.   
Additionally, when training a regression model, losses are proportional to the error magnitude, while for classification all errors receive an equal penalty: predicting bin 10 instead of 20, is just as incorrect as predicting bin 10 instead of bin 100. 
So classification cannot add anything new to regression; or can it?

Surprisingly, adding a classification loss to the regression loss \cite{mousavian20173d,workman2016horizon,xiao2019pose,zhou2019objects}, or even replacing the regression loss with classification \cite{ding2021classification, fu2018deep, tulsiani2015viewpoints} is extensively used in practice when training deep models for predicting continuous outputs.
The classification is typically defined by binning the regression targets into a fixed number of classes.
This is shown to be beneficial for tasks such as: depth estimation \cite{fu2018deep}, horizon line detection \cite{workman2016horizon}, object orientation estimation \cite{mousavian20173d,zhou2019objects}, age estimations \cite{rothe2015dex}.
And the reported motivation for discretizing the regression loss is that: 
it improves performance \cite{workman2016horizon,xiao2019pose}, 
or that it helps in dealing with noisy data \cite{workman2016horizon}, 
or that it helps overcome the overly-smooth regression predictions \cite{mousavian20173d,walker2015dense},
or that it helps better regularize the model \cite{kendall2017end, workman2016horizon}. 
However, none of these assumptions has been thoroughly investigated. 

In this work, we aim to explore in the context of deep learning: \emph{Why does classification help regression?} 
Intuitively, the regression targets contain more information than the classification labels. 
And adding a classification loss does not contribute any novel information.
What is it really that a classification loss can add to a standard MSE (mean squared error) regression loss? And why does it seem beneficial in practice? 

\begin{figure}
    \centering
    \includegraphics[width=.49\textwidth]{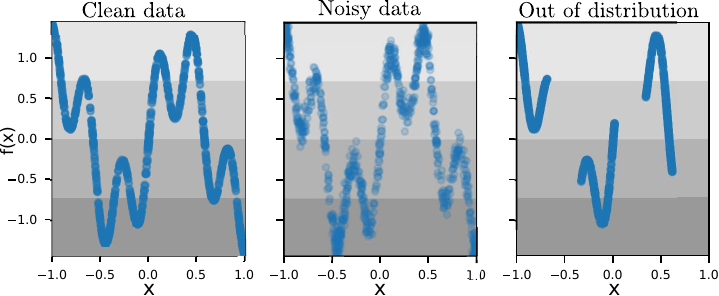}
    \caption{ 
            To probe \emph{``Why does classification help regression?"} we design a fully controlled dataset including the following scenarios:  
            \textsl{Clean data} -- 1$D$ non-linear functions defined by the sum of two sine waves with different frequencies and amplitudes; 
            \textsl{Noisy data} -- uniform noise added to the outputs; 
            \textsl{Out of distribution} -- sampling different regions of the input space during training and during testing. 
            (We show a single function here. The gray shading groups the function targets into 4 classes, as an example.)
    }
    \label{fig:cases}
\end{figure}

\begin{figure}[t]
    \centering
    \centerline{\includegraphics[width=.5\textwidth]{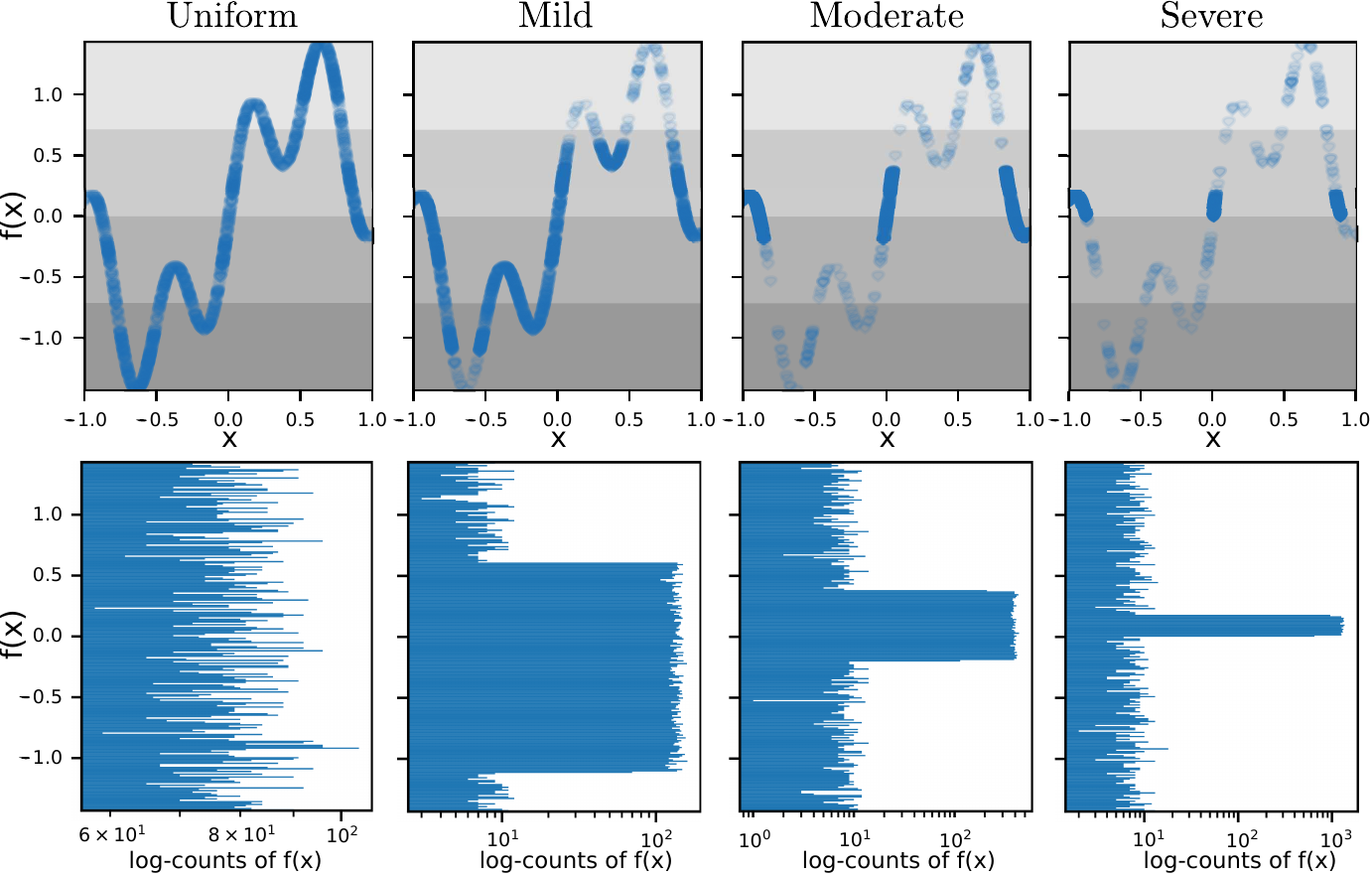}}
    \caption{ 
        Data sampling. 
        On the columns we increase the data imbalance from uniform (balanced) to severely imbalanced.
        On the first row we show the function $f(\mathbf{x})$ where darker datapoint colors visualize higher density. 
        The gray shading on the first row groups the targets into 4 classes.
        On the second row we show the log-counts per function value, sampled for the training data.
        We sample the test data uniformly. 
    }
    \label{fig:sampling}
\end{figure}

To take a step towards understanding why classification helps regression, we start the analysis in a fully controlled setup, using a set of 1$D$ synthetic functions.
We consider several prior hypothesis of when classification can help regression: \textsl{noisy} data, \textsl{out-of-distribution} data, and the normal \textsl{clean} data case, as in \fig{cases}.
Additionally, we vary the sampling of the data from uniform to highly imbalanced as in \fig{sampling}.
We empirically find out in which of these cases adding a classification loss improves the regression quality on the test set. 
Moreover, we explain these empirical observations by formulating them into a probabilistic analysis.

We urge the reader to note that our goal is not proposing a novel regression loss, nor do we aim to improve results with ``superior performance'' over state-of-the-art regression models.
But rather, we aim to investigate a common computer vision practice: \ie adding a classification loss to the regression loss, and we analyze for what kind of dataset properties and dataset samplings this practice is useful.
Finally, we show experimentally that our findings hold in practice on two imbalanced real-world computer vision datasets: NYUD2-DIR (depth estimation) and IMDB-WIKI-DIR (age estimation) \cite{yang2021delving}, and on the Breakfast dataset \cite{Kuehne12} when predicting video progression.

\section{Why does classification help regression?}
\label{sec:method}
For an input dataset, $\mathcal{D}$, containing N samples of the form $(\mathbf{x}, {y}){\in}\mathcal{D}$, we analyze what happens when we train a deep network with parameters $\boldsymbol{\omega}$ to predict a target ${y}^*{=}f(\mathbf{x}, \boldsymbol{\omega})$ for a sample $\mathbf{x}$, by minimizing NLL (negative log-likelihood) or equivalently, minimizing the MSE (mean squared error) regression loss: \begin{align}
    \boldsymbol{\omega}^* &= \argmin_{\boldsymbol{\omega}} \sum_{(\mathbf{x}, {y}){\in}\mathcal{D}} L({y},\mathbf{x},\boldsymbol{\omega}) \\
    &= \argmin_{\boldsymbol{\omega}} \left(\sum_{(\mathbf{x}, {y}){\in}\mathcal{D}} - \log p({y} |\mathbf{x},\boldsymbol{\omega}) \right) \\
    \label{eq:reg}
    &\equiv \argmin_{\boldsymbol{\omega}} \lambda \frac{1}{N}\sum_{(\mathbf{x}, {y}){\in}\mathcal{D}} \left( {y} - {y}^* \right)^2
\end{align}
where $\boldsymbol{\omega}^*$ are the optimal parameters, and we can reinterpret the imbalanced likelihood as the mean of a Gaussian distribution with $\sigma$ noise: $p({y}|\mathbf{x},\boldsymbol{\omega}) = \mathcal{N}({y}; {y}^*, \sigma^2 I)$, in which case minimizing the NLL is equivalent to minimizing the MSE loss \cite{bishop2006pattern}, and $\lambda$ is a function of the noise $\sigma$.

We contrast \eq{reg} to the case when we discretize the targets ${y}$ into a set of $C$ classes and use a classification loss next to the regression loss:
\begin{align}
\label{eq:reg_cls}
    L({y},\mathbf{x},\boldsymbol{\omega}) =& \lambda \left({y} - {y}^* \right)^2 - \log  p( {y}^*_c | \mathbf{x}, \boldsymbol{\omega}),
\end{align}
where $y^*_k$, $k{\in}\{1,..,C\}$ denotes the model predictions binned into classes, and specifically $y^*_c$ is the prediction at the true class indexed by $c$, for the sample $\mathbf{x}$. 
For the classification term, we make the standard \textsl{softmax} distribution assumption.

\subsection{Controlled 1D analysis}
\label{sec:oned}
To probe the question "\emph{Why does classification help regression?}" we first want to know in which cases does classification help regression.
We measure test-time MSE scores for each case in \fig{cases}, and compare training with a regression loss as in \eq{reg}, with training using an extra classification loss as in \eq{reg_cls}.
We randomly sample 10 functions of the form: $f(x){=}a \sin(cx) + b \sin(dx)$, where $f(x){\in}[-1.5, 1.5]$ and $x{\in}[-1,1]$. 
For every function we vary the dataset scenario as in \fig{cases}, and we also vary the sampling of the data from uniform to severely imbalanced sampling, as in \fig{sampling}.
Each dataset sampling is repeatedly performed with 5 different random seeds, where we always sample ${\approx}30,000$ samples in total, and then randomly pick 1/3 for training, for validation, and for testing, respectively.   
In the $\textsl{out-of-distribution}$ case, the training set misses certain function regions that are present in the validation/test set, and vice-versa; and there is an overlap of 1/4 between the function regions in the training set and the regions in the validation/test set.
For the imbalanced sampling, we aim to sample a range of the targets ${y}$ more frequently than other ranges. 
For this, we randomly select a location along the y-axis in each repetition: this defines the center of the peak in \fig{sampling}, second row. 
And depending on the sampling scenario, we use a fixed variance ratio around the peak (0.3, 0.1 and 0.03 for \textsl{mild}, \textsl{moderate} and \textsl{severe} sampling) to define the region from which we draw the frequent samples.
We sample 75\% of the samples from the peak region, and the rest uniformly from the other function areas.
\begin{figure}
    \centering
    \centerline{\includegraphics[width=0.5\textwidth]{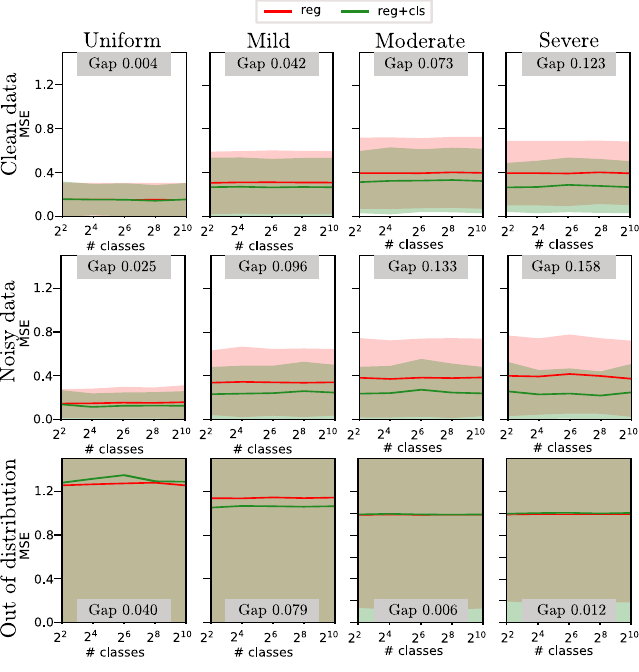}} 
    \caption{ 
        MSE per class, where we vary the number of classes from 4 to 1024. 
        We evaluate on uniformly sampled test sets, across 5 repetitions.
        We plot means and standard deviations for each dataset (rows) and sampling variation (columns), where the shading represents the standard deviation. 
        The red line, \textsl{reg}, should be constant across classes but it varies due to the sampling/training randomness.
        We also print the gap between the \textsl{reg} and \textsl{reg+cls} measured as absolute difference of MSE scores.
        The effect of the classification loss is present when the sampling of the data is imbalanced for the \textsl{clean} and \textsl{noisy} data.
    }
    \label{fig:1d_results}
\end{figure}

We train a simple MLP (multi layer perceptron) with 3 linear layers ($[1{\times}6]$, $[6{\times}16]$, $[16{\times}1]$) and ReLU non-linearities. 
For setting the hyperparameter $\lambda$, we perform a hyper-parameter search on the validation set.
We find the best $\lambda$ to be $1e{+}2$, $1e{+}3$ and $1e{+}4$ for the \textsl{clean}, \textsl{noisy} and \textsl{out-of-distribution}.
We train for 80 epochs using an Adam optimizer with a learning rate of $1e{-}3$, $1e{-}2$ and $1e{-}4$ for \textsl{clean}, \textsl{noisy} and \textsl{out-of-distribution} respectively.
We use a weight decay of $1e{-}3$. 
For classification we add at training-time a linear layer ($[16{\times }C]$) predicting $C$ classes. 
More details are in the supplementary material.

In \fig{1d_results} we show the MSE across all dataset and sampling variations, for $\{2^2, 2^4, 2^6, 2^8, 2^{10}\}$ classes.
We define the class ranges uniformly. 
The test sets are uniformly sampled and we perform 5 repetitions. 
We plot the means and standard deviations.
We print on every plot the gap between the \textsl{reg} and \textsl{reg+cls} measured as the average absolute difference of MSE scores.
From this 1$D$ analysis, we observe that the effect of the classification loss is visible when the training data is imbalanced for \textsl{clean} and \textsl{noisy} data. 

\subsection{Anchoring 1$D$ experimental observations}
In \sect{oned} we observe that classification has a more pronounced effect when the sampling of the data is imbalanced.
Therefore, from here on we focus the analysis on imbalanced data sampling.
We start from the derivations of Ren \etal \cite{ren2022balanced} who define the relation between the NLL (negative log-likelihood) of  imbalanced samples $- \log \widetilde{p}({y}| \mathbf{x},\boldsymbol{\omega})$ and the NLL of the balanced samples $- \log p({y}| \mathbf{x},\boldsymbol{\omega})$:
\begin{align}
\label{eq:imbalanced}        
    -\log \widetilde{p}({y}|\mathbf{x},\boldsymbol{\omega}) 
        &= - \log \frac{p({y}|\mathbf{x},\boldsymbol{\omega}) \widetilde{p}({y})}{\int_{{y}^\prime} p({y}^\prime | \mathbf{x},\boldsymbol{\omega}) \widetilde{p}({y}^\prime) d{y}^\prime} 
\end{align}
where $\widetilde{p}({y})$ denotes the prior over imbalanced targets.
\eq{imbalanced} holds under the assumption that the data function remains unchanged, $\widetilde{p}(\mathbf{x} | y){=}p(\mathbf{x} | y)$, which is the case for the \textsl{clean} and \textsl{noisy} data scenarios above.
We decompose the $\log$ and rewrite the relation between the NLL of the balanced and the NLL of the imbalanced data:
\begin{align}
\label{eq:nll-extra}        
    -\log \widetilde{p}({y}| \mathbf{x},\boldsymbol{\omega}) + L_\text{extra}({y},\mathbf{x},\boldsymbol{\omega}) = & - \log p({y}|\mathbf{x},\boldsymbol{\omega}),
\end{align} 
$L_\text{extra}$ contains all the information about the imbalanced regression targets:
\begin{align}
    L_\text{extra} = & \log \widetilde{p}({y}){-}\log \int_{{y}^\prime} p({y}^\prime|\mathbf{x},\boldsymbol{\omega}) \widetilde{p}({y}^\prime)  d{y}^\prime,\\ 
   = & \log \widetilde{p}({y}){-}\log \int_{{y}^\prime} \mathcal{N}({y}^\prime; {y}^*, \sigma^2 I)\widetilde{p}({y}^\prime) d{y}^\prime,
\end{align} 
where again ${y}^*$ are the predicted targets. 

To derive the link between optimizing a model on imbalanced data and using both a regression MSE loss and a classification loss, we assume the imbalanced regression targets ${y}$ can be discretized into a set of classes, $k{\in}\{1, .., C\}$ such that $\sum_{k=1}^C {p}({y}_k){=}1$.
By going from continuous regression targets to discrete classes, we change the form of the log-likelihood from Gaussian to \textsl{softmax}:
\begin{align}
\label{eq:discrete}
    L_\text{extra} &\approx \log \widetilde{p}({y}_c) -\log \sum_{k=1}^C p({y}_{k}^* | \mathbf{x}, \boldsymbol{\omega}) \widetilde{p}({y}_k),
\end{align} 
where we denote the true class label by $y_c$.
Note that the regression targets $y$ are imbalanced, but the classes ${y}_{k}$ do not necessarily need to be imbalanced.
We analyze in the experimental section the effect of defining balanced classes.

We make the observation that if we could optimize the class assignment, the $L_\text{extra}$ term would disappear.
If the classes are optimized, then the class likelihoods are close to 0 for all classes except the true class: $p({y}^*_k | \mathbf{x}, \boldsymbol{\omega}){\approx}0$, $\forall k{\neq}c$, where $c$ indexes the true class.
Using this in the expression of $L_\text{extra}$, we obtain:
\begin{align}
    L_\text{extra} \approx & \log \widetilde{p}({y}_c) - \log p({y}_{c}^* | \mathbf{x}, \boldsymbol{\omega}) \widetilde{p}({y}_c), \\
    \approx & - \log p({y}_{c}^* | \mathbf{x}, \boldsymbol{\omega}),
\end{align}     
where the $\widetilde{p}({y}_c)$ terms cancel out when decomposing the second $\log$.
Therefore, we can see that optimizing the class cross-entropy loss reduces the gap between between the NLL of imbalanced data and NLL of balanced data.
(Note: we observe in practice that if the classifier fails to converge, adding a classification loss is detrimental to regression.)

\subsection{Defining balanced classes in practice}
\label{sec:bal}
Existing works show that optimizing imbalanced classes is problematic \cite{li2020overcoming,zhang2023deep}.
In practice, researchers opt for using balanced classes in combination with regression \cite{mousavian20173d,workman2016horizon,xiao2019pose,zhou2019objects}. 
Here, the data is imbalanced, however we are free to define the class ranges such that we obtain balanced classes over imbalanced data sampling.
To empirically test the added value of using balanced classes, we need a way to define balanced classes over imbalanced data sampling.

Given an imbalanced data sampling, we bin samples into classes, using uniform class ranges. 
This generates imbalanced classes, which we then re-balance by redefining the class ranges such that the class histogram is approximately uniform.
To this end, we apply histogram equalization over the original classes:
\begin{align}
    \label{eq:balance_cls}
    q(k) = \left\lfloor\frac{C}{N} \sum_{j=1}^k \mathcal{H}_C(j)\right\rfloor, 
\end{align}
where $\lfloor x \rfloor$ rounds $x$ down to the nearest integer,
and $\mathcal{H}_C(\cdot)$ computes the histogram of the samples per class,
and $q(\cdot)$ is a mapping function that maps the old classes indexed by $k{\in}\{1,..,C\}$ to a new set of classes $\{1,..,\overline{C}\}$.
\Eq{balance_cls} merges class ranges such that their counts are as close as possible. 
Thus, the number of equalized classes is lower or equal to the original number of classes, $\overline{C} \le C$. 

After class equalization, the new classes are not perfectly uniform. 
We further define a class-keeping probability $\rho(k)$, as the ratio between the minimum class count and the current equalized class count $\mathcal{H}_{\overline{C}}(k)$:
\begin{align}
 \label{eq:bal_keep}
 \rho(k) = \frac{\min_{j=1}^{\overline{C}} \mathcal{H}_{\overline{C}}(j)}{\mathcal{H}_{\overline{C}}(k)},
\end{align}
where $\mathcal{H}_{\overline{C}}(\cdot)$ computes the histogram of equalized classes. 
Selecting training samples using only \eq{bal_keep}, without first equalizing the classes, will lead to never seeing samples from the most frequent classes.
During training, for the regression loss we use all samples, while for the classification loss we pick samples $(\mathbf{x},y_k)$ with a probability defined by $\rho(k)$. More details are in the supplementary material.
 
\section{Empirical analysis}

\begin{figure}
    \centering
    \centerline{\includegraphics[width=.5\textwidth]{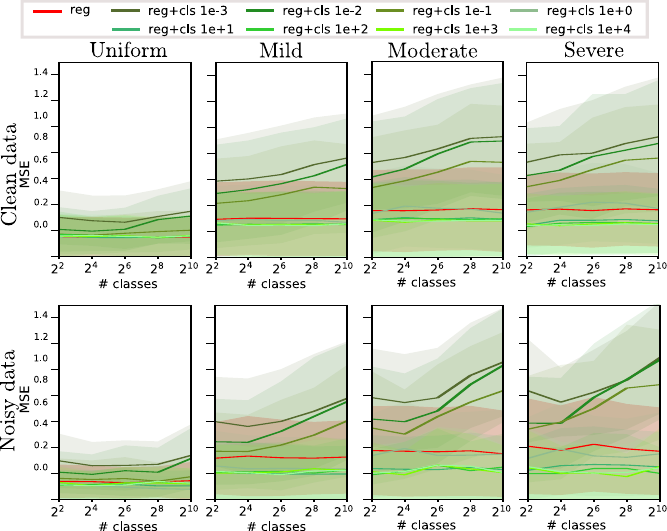}}
    \caption{
        \textbf{Effect of the $\lambda$ hyperparameter on the validation:} 
        We evaluate a range of values $\lambda{\in}\{1e{-}3, 1e{-}2, 1e{-}1, 1, 1e{+}1,$ $ 1e{+}2, 1e{+}3, 1e{+}4\}$ on the validation set.
        The shading represents the standard deviation of the MSE error across 3 repetitions.
        Setting $\lambda$ correctly is essential when using a classification loss next to the regression loss.
    }
    \label{fig:lambda}
\end{figure}

\subsection{Hypothesis analysis on 1D data}
We use the 10 randomly sampled 1$D$ functions to further analyze the regression loss --- \textsl{reg} from \eq{reg}, and regression with classification --- \textsl{reg${+}$cls} from \eq{reg_cls}.
We start with the 1$D$ data because it is easily interpretable and it offers a controlled environment to test the hypothesis that classification helps regression and to analyze the properties of the classes.
\footnote{Our source code will be made available online, at the address: \href{https://github.com/SilviaLauraPintea/reg-cls}{https://github.com/SilviaLauraPintea/reg-cls}.}

\smallskip\noindent\textbf{Effect of the $\lambda$ hyperparameter}.
We perform hyperparameter search on the validation set for setting the $\lambda$ in \eq{reg_cls}.
We vary $\lambda{\in}\{1e{-}3, 1e{-}2, 1e{-}1, 1, 1e{+}1, 1e{+}2,$ $ 1e{+}3, 1e{+}4\}$.
\fig{lambda} shows the MSE across 3 repetitions, when considering different number of classes, for \textsl{clean} and \textsl{noisy} data scenarios, across sampling variations.
Higher values of $\lambda$ typically perform better in this case.
When the sampling of the data is imbalanced, there exists a value of $\lambda$ such that the \textsl{reg} baseline is outperformed by the \textsl{reg+cls}. 
We use the best $\lambda$ values found on the validation, when evaluating on the test set.

\smallskip\noindent\textbf{Effect of balancing the classes on imbalanced data.}
We numerically evaluate in \tab{balanced} if using balanced classes (as defined in \Eq{balance_cls}-\Eq{bal_keep}) is less sensitive to the choice of $\lambda$.
We perform 3 repetitions over all dataset scenarios and sampling cases, and vary $\lambda{\in}\{1e{-}3, 1e{-}2,$ $ 1e{-}1, 1, 1e{+}1, 1e{+}2, 1e{+}3, 1e{+}4\}$.
For this we consider the percentage of runs (across different number of classes, random seeds, and values of $\lambda$) where classification helps regression --- where \textsl{reg+cls} MSE is lower than \textsl{reg} MSE.
Ideally this number should be close to 100\%. 
Balancing the classes is more robust to the choice of $\lambda$, as on average there are more runs where classification helps regression.
\begin{table}
    \centering
    \resizebox{.9\linewidth}{!}{
    \begin{tabular}{l cc cc}
        \toprule
        & \multicolumn{2}{c}{Imbalanced classes ($\uparrow$)} & \multicolumn{2}{c}{Balanced classes ($\uparrow$)}\\ \cmidrule(lr){2-5}
        Dataset case  & Clean & Noisy data & Clean & Noisy data \\ \cmidrule(lr){2-3}\cmidrule(lr){4-5}
        Uniform       & 55.42\% & 59.83\% &  59.00\% & 65.00\% \\
        Mild          & 54.17\% & 55.75\% &  57.67\% & 62.42\%  \\
        Moderate      & 47.50\% & 56.58\% &  47.50\% & 56.00\%  \\
        Severe        & 36.83\% & 49.08\% &  34.25\% & 46.25\%  \\ \midrule 
        Avg           & 48.48\% & 55.31\% &  49.60\% & 57.42\%  \\ \bottomrule
    \end{tabular}}
        \caption{
        \textbf{Effect of balancing the classes:} 
        On the validation sets we test how sensitive \textsl{reg+cls} is to the choice of $\lambda$ when balancing classes versus when using imbalanced classes.
        For this we vary $\lambda{\in}\{1e{-}3, 1e{-}2, 1e{-}1, 1, 1e{+}1, 1e{+}2, 1e{+}3, 1e{+}4\}$.
        And we measure the percentage of runs where adding a classification loss improves the regression predictions.
        Using balanced classes is less sensitive to the choice of $\lambda$.
    }
    \label{tab:balanced}
\end{table}

\begin{figure}
    \centering
    \centerline{\includegraphics[width=.5\textwidth]{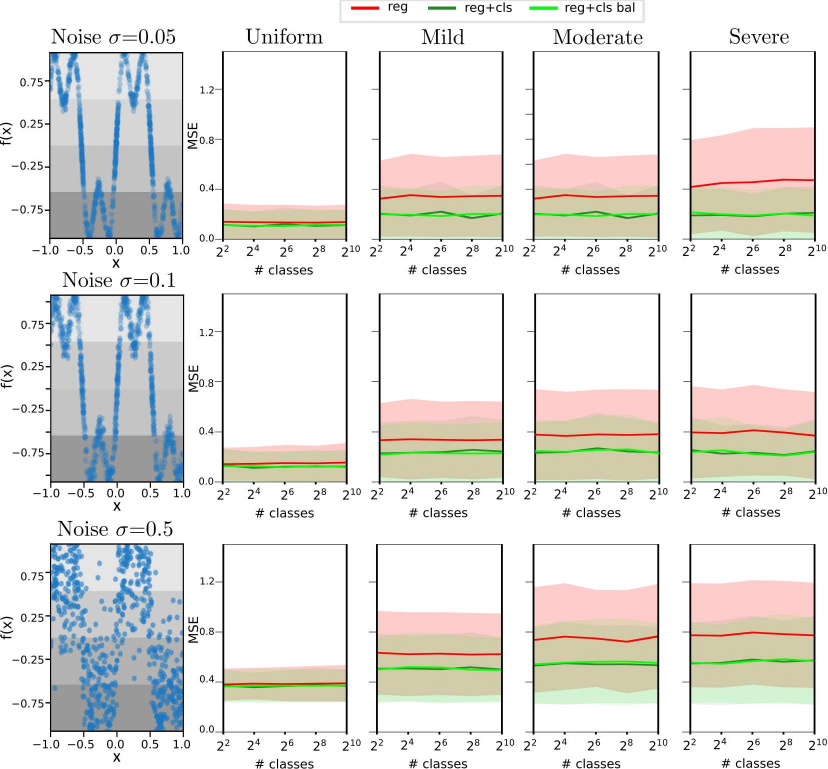}}
    \caption{
        \textbf{Effect of noisy targets:} 
        We vary the amount of noise in the targets, along the y-axis.
        We plot both using imbalanced classes (green), and using balanced classes (lime) compared to the \textsl{reg} baseline (red).
        We report MSE on the test sets across 5 repetitions.
        Despite the the noise level increasing drastically, the benefits of adding a classification loss to the regression loss remain visible.
    }
    \label{fig:noise_level}
\end{figure}

\smallskip\noindent\textbf{Effect of the noisy targets.}
In \fig{1d_results} we considered a single noise level for the \textsl{Noisy data} scenario.
Here, we further analyze the effect on the MSE scores when varying the noise level in the targets for the baseline \textsl{reg} (in red) and the \textsl{reg+cls} with imbalanced classes (in green) as well as \textsl{reg+cls bal} where the classes are balanced (in lime color).
We vary the level of noise $\sigma$ of the targets on the y-axis $\sigma{\in}\{0.05, 0.1, 0.5\}$, and plot mean MSE and standard deviation on uniform test sets across 5 repetitions. 
For the balanced case, we indicate the original number of classes on the x-axis, while in practice the number of balanced classes is typically $2{\times}$ lower than the initial one.
\fig{noise_level} shows that despite severely increasing the noise level, adding a classification loss remains beneficial when the data is imbalanced. 

\begin{table*}
    \centering
    \resizebox{1.\linewidth}{!}{
    \begin{tabular}{rllllllll}
        \toprule
        & \multicolumn{4}{c}{NYUD2-DIR RMSE-val ($\downarrow$)} & \multicolumn{4}{c}{NYUD2-DIR RMSE-test ($\downarrow$)}\\ \cmidrule(lr){2-5}\cmidrule(lr){6-9}
        Samples                                       & All   & Many  & Med.  & Few & All   & Many  & Med.  & Few \\ \midrule
        Kernel \cite{yang2021delving} & --- & --- & --- & --- & 1.338 & 0.670 & 0.851 & 1.880 \\
        Balanced \cite{ren2022balanced} & --- & --- & --- & --- & 1.251 & 0.692 & 0.959 & 1.703 \\
        \textsl{reg} (MSE) & 1.614 (${\pm}$0.051) & 0.554 (${\pm}$0.002) & 0.934 (${\pm}$0.042) & 2.360 (${\pm}$0.081) &
        1.499 (${\pm}$0.083) & 0.578 (${\pm}$0.010) & 0.896 (${\pm}$0.043) & 2.171 (${\pm}$0.136)\\ \midrule   
        \textsl{reg${+}$cls} (2 cls) & 1.587 (${\pm}$0.026) & 0.618 (${\pm}$0.003) & 1.062 (${\pm}$0.025) & 2.278 (${\pm}$0.041) & 1.532 (${\pm}$0.082) & 0.624 (${\pm}$0.036) & 0.946 (${\pm}$0.032) & 2.204 (${\pm}$0.141) \\
        \textsl{reg${+}$cls} (10 cls) & 1.576 (${\pm}$0.063) & 0.585 (${\pm}$0.008) & 0.982 (${\pm}$0.058) & 2.282 (${\pm}$0.095)  & 1.509 (${\pm}$0.022)  & 0.582 (${\pm}$0.013) & 0.947 (${\pm}$0.046) & 2.178 (${\pm}$0.047)\\
        \textsl{reg${+}$cls} (100 cls)\vspace{2px} & 1.536 (${\pm}$0.090) & 0.569 (${\pm}$0.018) & 0.966 (${\pm}$0.041) & 2.222 (${\pm}$0.146) & 1.488 (${\pm}$0.028) & 0.578 (${\pm}$0.015) & 0.971 (${\pm}$0.049) & 2.141 (${\pm}$0.045) \\
        \cdashline{2-9}\\[-5px]
        \textsl{reg${+}$cls bal.} (2 cls)   & 1.599 (${\pm}$0.020) &  0.616 (${\pm}$0.026) & 1.033 (${\pm}$0.058) & 2.304 (${\pm}$0.044) & 1.522 (${\pm}$0.060) & 0.665 (${\pm}$0.033) & 1.003 (${\pm}$0.059) & 2.166 (${\pm}$0.118) \\
        \textsl{reg${+}$cls bal.} (10 cls)  & 1.454 (${\pm}$0.044) & 0.607 (${\pm}$0.041) & 0.965 (${\pm}$0.023) & 2.077 (${\pm}$0.087) & 1.454 (${\pm}$0.044) & 0.607 (${\pm}$0.041) & 0.965 (${\pm}$0.023) & 2.077 (${\pm}$0.087) \\
        \textsl{reg${+}$cls bal.} (100 cls) & 1.553 (${\pm}$0.117) & 0.563 (${\pm}$0.033) & 0.897 (${\pm}$0.043) & 2.263 (${\pm}$0.193) & 1.487 (${\pm}$0.051) & 0.574 (${\pm}$0.014) & 0.869 (${\pm}$0.019) & 2.156 (${\pm}$0.084) \\ \bottomrule
    \end{tabular}}
    \caption{\small 
        \textbf{Imbalanced realistic image data: NYUD2-DIR depth estimation.}
        We evaluate the baseline \textsl{reg} trained with MSE, and the \textsl{reg${+}$cls} variants.
        We report RMSE when the best model is selected on the validation (\textsl{RMSE-val}), or as in \cite{yang2021delving} on the test set (\textsl{RMSE-test}).
        We average over 3 different random seeds.
        Adding a classification loss helps regression, which validates our hypothesis.
        }
\label{tab:nyud2}
\end{table*}

\begin{table*}[!h]
    \centering
    \resizebox{1.\linewidth}{!}{
    \begin{tabular}{rllllllll}
        \toprule
        & \multicolumn{4}{c}{IMDB-WIKI-DIR MAE($\downarrow$)} & \multicolumn{4}{c}{IMDB-WIKI-DIR MSE($\downarrow$)}\\ \cmidrule(lr){2-5}\cmidrule(lr){6-9}
        Samples                                        & All & Many  & Med. & Few & All & Many & Med. & Few \\ \midrule
        Focal \cite{lin2017focal}                 & 7.97 & 7.12 & 15.14 & 26.96 & 136.98 & 106.87 & 368.60 & 1002.90 \\
        Kernel \cite{yang2021delving}             & 7.78 & 7.20 & 12.61 & 22.19 & 129.35 & 106.52 & 311.49 & 811.82 \\ 
        Balanced \cite{ren2022balanced}           & 8.12 & 7.58 & 12.27 & 23.05 & --- & --- & --- & --- \\
        \textsl{reg} (MAE)                        & 8.09 (${\pm}$0.01) & 7.23 (${\pm}$0.02) & 15.48 (${\pm}$0.15) & 26.81 (${\pm}$0.48) & 138.53 (${\pm}$1.17) & 107.82 (${\pm}$0.96) & 375.27 (${\pm}$6.97) & 1017.59 (${\pm}$27.51) \\ \midrule 
        \textsl{reg${+}$cls} (2 cls)  & 7.95 (${\pm}$0.05) & 7.11 (${\pm}$0.03) & 15.08 (${\pm}$0.27) & 26.15 (${\pm}$0.06) & 135.15 (${\pm}$0.40) & 105.86 (${\pm}$0.27) & 361.02 (${\pm}$7.00) & 973.53 (${\pm}$11.52) \\
        \textsl{reg${+}$cls} (10 cls) & 7.93 (${\pm}$0.06) & 7.12 (${\pm}$0.06) & 14.93 (${\pm}$0.17) & 25.91 (${\pm}$0.27) & 135.69 (${\pm}$1.65) & 106.58 (${\pm}$1.42) & 359.27 (${\pm}$5.70) & 975.37 (${\pm}$34.09) \\
        \textsl{reg${+}$cls} (100 cls) \vspace{2px} & 7.61 (${\pm}$0.02) & 6.90 (${\pm}$0.03) & 13.46 (${\pm}$0.44) & 25.04 (${\pm}$ 0.82) & 129.73 (${\pm}$1.33) & 103.25 (${\pm}$1.12) & 328.17 (${\pm}$ 10.40) & 933.33 (${\pm}$ 9.16) \\\cdashline{2-9}\\[-5px]
        \textsl{reg${+}$cls bal.} (2 cls)   & 7.94 (${\pm}$0.06) & 7.14 (${\pm}$0.09) & 14.71 (${\pm}$0.51) & 26.20 (${\pm}$0.73) & 134.91 (${\pm}$1.20) & 106.36 (${\pm}$1.68) & 351.48 (${\pm}$14.82) & 980.61 (${\pm}$39.44) \\
        \textsl{reg${+}$cls bal.} (10 cls)  & 7.92 (${\pm}$0.06) & 7.10 (${\pm}$0.05) & 14.99 (${\pm}$0.21) & 25.58 (${\pm}$0.39) & 134.53 (${\pm}$0.75) & 105.40 (${\pm}$0.43) & 362.48 (${\pm}$4.41) & 940.91 (${\pm}$18.70) \\
        \textsl{reg${+}$cls bal.} (100 cls) & 7.59 (${\pm}$0.09) & 6.86 (${\pm}$0.08) & 13.61 (${\pm}$0.35) & 25.30 (${\pm}$0.19) & 127.87 (${\pm}$1.99) & 101.34 (${\pm}$1.74) & 327.55 (${\pm}$12.98) & 925.65 (${\pm}$21.96) \\
        \bottomrule
    \end{tabular}}
    \caption{\textbf{Imbalanced realistic image data: IMDB-WIKI-DIR age estimation.} 
        We evaluate the baseline \textsl{reg} when trained with MAE, and the \textsl{reg${+}$cls} variants.
        We report test MSE and MAE on the test set, averaged over 3 repetitions with different random seeds.
        Adding a classification loss next to the regression loss is specifically beneficial on this dataset, where this simple yet popular strategy performs on par with state-of-the-art.
        (We use ``---" where the authors' results are missing).
    }
    \label{tab:imdbwiki}.
\end{table*}
\begin{table}
    \centering
    \resizebox{.7\linewidth}{!}{
    \begin{tabular}{rll}
        \toprule
        & \multicolumn{2}{c}{Balanced NYUD2-DIR} \\ \cmidrule(lr){1-3}
        & All RMSE-val ($\downarrow$)  & All RMSE-test ($\downarrow$) \\ \midrule
        \textsl{reg} (MSE)   & 1.442 (${\pm}$0.077) & 1.492 (${\pm}$0.042) \\ 
        \textsl{reg${+}$cls} & 1.456 (${\pm}$0.033) & 1.593 (${\pm}$0.025) \\ \midrule
        & \multicolumn{2}{c}{Balanced IMDB-WIKI-DIR} \\ \cmidrule(lr){1-3}
        & All MAE ($\downarrow$)  & All MSE ($\downarrow$) \\ \midrule
        \textsl{reg} (MSE)   & 7.74 (${\pm}$0.04) & 131.03 (${\pm}$1.44) \\ 
        \textsl{reg${+}$cls} & 7.71 (${\pm}$0.06) & 131.27 (${\pm}$1.00) \\ 
        \bottomrule
    \end{tabular}}
    \caption{\small 
        \textbf{Balanced realistic image data: NYUD2-DIR depth estimation and IMDB-WIKI-DIR age estimation.}
        We compare the \textsl{reg} and \textsl{reg+cls} (using 100 classes) when the data is balanced. We report RMSE and MAE\slash MSE, respectively, across 3 repetitions.
        There are no clear improvements when adding a classification loss to the regression on balanced data.
        }        
\label{tab:bal}
\end{table}

\begin{table*}[t]
    \centering
    \resizebox{1.\linewidth}{!}{
    \begin{tabular}{r llll llll}
        \toprule
        & \multicolumn{4}{c}{Breakfast RMSE \% ($\downarrow$)} & \multicolumn{4}{c}{Breakfast RMSE frames ($\downarrow$) -- unnormalized}\\ \cmidrule(lr){2-5}\cmidrule(lr){6-9}
        Dataset split                  & S1  & S2  & S3  & S4 & S1 & S2 & S3  & S4 \\ \midrule
        Random baseline                & 58.50 (${\pm}$0.29) & 58.37 (${\pm}$0.07) & 58.39 (${\pm}$0.14) & 58.38 (${\pm}$0.12) & 1394.48 (${\pm}$992.50) & 1511.04 (${\pm}$1132.60) & 1450.26 (${\pm}$1085.56) & 1420.21 (${\pm}$1063.16)\\
        \cite{kukleva2019unsupervised} & 31.24 (${\pm}$0.80) & 31.49 (${\pm}$0.98) & 30.85 (${\pm}$0.66) & 30.78 (${\pm}$0.53) & 1079.38 (${\pm}$775.17) & 1235.02 (${\pm}$932.89) & 1172.67 (${\pm}$880.69) & 1170.07 (${\pm}$886.92)\\ 
        \textsl{reg} (RMSE)            & 32.57 (${\pm}$1.45) & 33.30 (${\pm}$1.69) & 32.52 (${\pm}$1.17) & 33.08 (${\pm}$1.49) & 860.17 (${\pm}$583.44) & 891.39 (${\pm}$641.58) & 862.65 (${\pm}$609.01) & 845.48 (${\pm}$605.46)\\ \midrule  
        \textsl{reg${+}$cls} (100 cls) & 28.71 (${\pm}$0.38) & 28.84 (${\pm}$0.55) & 28.46 (${\pm}$0.48) & 28.44 (${\pm}$0.50) & 837.59 (${\pm}$573.15) & 870.55 (${\pm}$630.89) & 845.65 (${\pm}$601.73) & 809.76 (${\pm}$595.37)\\ 
        \textsl{reg${+}$cls bal.} (100 cls) & ---  & ---  & --- & --- & 837.61 (${\pm}$573.17) & 870.54 (${\pm}$630.88) & 845.65 (${\pm}$601.72) & 809.76 (${\pm}$595.37) \\ \bottomrule
        & \multicolumn{4}{c}{(a) Percentage prediction (normalized).} 
        & \multicolumn{4}{c}{(b) Frame prediction (unnormalized).}\\
    \end{tabular}}
    \caption{\textbf{Imbalanced video progress prediction on Breakfast.}
         We report mean RMSE and standard deviations averaged over all 10 cooking tasks of the Breakfast dataset.
        (a) Progress prediction in video percentages.
        (b) Progress prediction in frame numbers.
        The overall progress prediction results leave space for improvements for all methods, because of the dataset challenges.
        However, also for this regression problem adding a classification loss has benefits.
    }
    \label{tab:bf}
\end{table*}

\subsection{Realistic image datasets}
\footnotetext[1]{Using more epochs on \textsl{NYUD2-DIR} seems to lead to overfitting.}
\noindent\textbf{Imbalanced realistic image datasets.} We run experiments on two realistic imbalanced datasets from Yang \etal \cite{yang2021delving}: depth estimation on the \textsl{NYUD2-DIR} dataset, and age estimation on the \textsl{IMDB-WIKI-DIR}.
The supplementary material plots dataset statistics.
For both datasets we use the Adam optimizer and set the learning rate to $1e{-}4$ for \textsl{NYUD2-DIR} with 5 epochs\protect\footnotemark[1] and batch size 16 accumulated over 2 batches (to mimic batch size 32 on 2 GPUs), while for \textsl{IMDB-WIKI} we use a learning rate of $1e{-}3$ for 90 epochs and batch size 128.
When comparing with Ren \etal \cite{ren2022balanced} we use their best results (their GAI method). 
We also use the evaluation code provided in Yang \etal \cite{yang2021delving} and we report RMSE (root mean squared error) on \textsl{NYUD2-DIR} and MAE (mean absolute error)\slash MSE (mean squared error) for \textsl{IMDB-WIKI-DIR} as also done in \cite{ren2022balanced,yang2021delving}.
When re-running the baseline \textsl{reg} results we observed a large variability across different runs by varying the random seed, especially on the \textsl{NYUD2-DIR} dataset, therefore we report results averaged over 3 random seeds.  
We use the architecture of Yang \etal \cite{yang2021delving}: ResNet-50 \cite{he2016deep} for \textsl{IMDB-WIKI-DIR}, and the ResNet-50-based model from \cite{hu2019revisiting} for \textsl{NYUD2-DIR}.

For adding the classification loss on \textsl{IMDB-WIKI-DIR} we append, only during training, a linear layer of size $[F, C]$ followed by a softmax activation and a cross-entropy loss, where $F$ is the number of channels in the one-to-last layer.
For the \textsl{NYUD2-DIR} the predictions are per-pixel, thus we use the segmentation head from Mask R-CNN \cite{he2017mask} composed of a transposed convolution of size $2{\times}2$, ReLU, and a $1{\times}1$ convolution predicting the number of classes, $C$.  
At test time the classification branch is not used.
We estimate the $\lambda$ hyperparameter using the validation set provided in \cite{yang2021delving} on \textsl{IMDB-WIKI-DIR}. 
For \textsl{NYUD2-DIR} we define a validation set by randomly selecting 1\slash 5 of the training directories with a seed of 0, and we use the same training\slash validation\slash test split for all our results. 
For \textsl{NYUD2-DIR} we use $\lambda{=}1.0$ for 100 classes and $\lambda{=}0.1$ for 10 and 2 classes. 
For \textsl{IMDB-WIKI-DIR} we set $\lambda{=}0.1$ for 100 classes and $\lambda{=}1.0$ for 10 and 2 classes. 
We compare the \textsl{reg} results with \textsl{reg${+}$cls} adding the classification loss, and  \textsl{reg${+}$cls bal.} with balanced classes, where we consider 2, 10 and 100 classes.

\tab{nyud2} shows the RMSE results on \textsl{NYUD2-DIR} when training with the standard MSE loss compared to adding a classification loss.
We report RMSE on the test, where the best model is selected on the validation set across epochs --- \textsl{RMSE-val}. 
To compare with previous work who selects the best model on the test, we also report this as \textsl{RMSE-test} (despite this being a bad practice). 
Additionally, note that our training set is slightly smaller because of using a validation set, so the \textsl{reg} results are worse than in \cite{yang2021delving} (\ie $1.477$ \textsl{RMSE-test}).
We observe that there is an inconsistency between the training and test set, as the best model on the test set does not correspond to the best model on the validation set (which is a subset of the training).
Despite all these, adding a classification loss still improves across all class-options, when selecting the best model on the validation set, and for 100 classes and 10 balanced classes, when selecting the best model on the test set.
This may be due to the classifier overfitting on the training data for fewer classes.

\tab{imdbwiki} gives the MAE and MSE results on \textsl{IMDB-WIKI-DIR} when using the standard MSE loss during training compared to when adding the classification at training time.
The best results are obtained using 100 balanced classes.
Here, adding a classification loss not only improves over the regression baseline, but it is also on-par with state-of-the-art methods specifically designed for imbalanced regression, such as \cite{ren2015faster,yang2021delving}. 
Adding a classification loss is similar to \cite{ren2015faster,yang2021delving}, who define smooth classes over the data. 

\medskip\noindent\textbf{Balanced realistic image datasets.}
On the same two datasets: \textsl{NYUD2-DIR} and \textsl{IMDB-WIKI}, we test the effect of adding a classification loss when we re-balance the data by binning the targets into 100 bins and selecting samples per batch during training as defined as in \Eq{balance_cls}-\Eq{bal_keep} for both \textsl{reg} and \textsl{reg+cls}. 
Because we mask samples per batch to balance the training data, for \textsl{IMDB-WIKI} here we use a batch size of 128 and learning rate $1e{-}4$ for 90 epochs.
While for \textsl{NYUD2-DIR} we use a learning rate of $1e{-}4$ and batch size of 8, accumulated over 4 batches (to mimic a batch of 32 on 1 GPU), for 5 epochs.
\tab{bal} shows the results across 3 repetitions.  
When the data is already balanced, adding a classification loss has limited effect. 

\subsection{Imbalanced video progress prediction}
As an additional investigation on imbalanced data, we explore videos which are naturally imbalanced in the number of frames. We perform video progress prediction on the Breakfast video dataset \cite{Kuehne12} containing 52 participants performing 10 cooking activities.
We follow the standard dataset split (S1, S2, S3, S4) \cite{Kuehne12} and use the corresponding train/test splits.
We adopt the method in \cite{kukleva2019unsupervised} and train a simple MLP on top of IDT (improved dense trajectory) features \cite{wang2013action} of dimension 64 over trajectories of 15 frames.
We evaluate RMSE when predicting either video progress percentages, or absolute frame numbers.
Kukleva \etal \cite{kukleva2019unsupervised} use an MLP with 3 linear layers and sigmoid activations.
We change the sigmoid activations into ReLU activations for the \textsl{reg} and \textsl{reg${+}$cls} models, since it works better when predicting absolute frame numbers.
For all methods we keep the training hyperparameters from \cite{kukleva2019unsupervised}: learning rate $1e{-}3$, Adam optimizer, and training for 40 epochs with the learning rate decreased by 0.1 at 30 epochs.
We also report the random baseline results when using the untrained model.
The data is imbalanced, in the sense that video lengths vary widely (see supplementary material for data distribution).
We test again if adding a classification loss can benefit the regression predictions when using 100 classes in the \textsl{reg${+}$cls}.
We search for $\lambda$ on a validation set created by randomly selecting 1/3 of the training videos with a seed of 0.
We use $\lambda{=}100$ when predicting percentages, and $\lambda{=}10$ when predicting frame numbers.
At training time, we add the classification loss via a linear layer on top of the one-to-last layer, followed by \textsl{softmax}.
We train one model per task and report mean and standard deviations over all 10 tasks.

\tab{bf} depicts the results across all 4 splits.
We report RMSE scores when predicting video progress in terms of percentages in \tab{bf}(a), and when predicting video progress in terms of frames in \tab{bf}(b).
Even if we predict video progress percentages between [0,100]\% in \tab{bf}(a), because the video length varies widely, for some videos we will have a lot more frame than for others, causing the data sampling to still be imbalanced.
In both cases there is gain from adding a classification loss to the regression loss. 

\section{Discussion and limitations}

\smallskip \noindent \textbf{Relation to nonlinear ICA.}
Hyv\"{a}rinen \etal \cite{hyvarinen2016unsupervised} show that there is a relation between learning to bin a continuous function and performing nonlinear ICA.
Hyv\"{a}rinen \etal \cite{hyvarinen2016unsupervised} start from a continuous signal $\mathbf{x}$ generated by a non-linear combination of source-signals: $\mathbf{x}{=}f(\mathbf{s})$, where $f(\cdot)$ is a non-linear function and $\mathbf{s}$ are the independent and non-stationary sources.
They split the signal into $C$ temporal segments $\mathbf{x}{=}\{\mathbf{x}_1, \mathbf{x}_2,.. ,\mathbf{x}_C\}$ and train an MLP to predict for every sample $\mathbf{x}_{k_i}$ the segment it belongs to: $g(\mathbf{x}_{k_i}, \theta){=}k, k{\in}\{1,..,C\}\text{ and }\mathbf{x}_{k_i}{\subset}\mathbf{x}_k,\text{ and }\theta$ are the MLP parameters.
Hyv\"{a}rinen \etal \cite{hyvarinen2016unsupervised} prove that the last hidden layer of the MLP $h_g(\cdot,\theta)$ recovers the original sources $\mathbf{s}$ within a linear transformation: $h_g(\mathbf{x}_{k_i},\theta) = \mathbf{w} \mathbf{s}_{k_i} + \mathbf{b}$, where $\mathbf{w},\mathbf{b}$ define a linear transformation.  
Intuitively, Hyv\"{a}rinen \etal \cite{hyvarinen2016unsupervised} discretize the signal into segments and classify the segments, thus performing classification on a continuous function. 
However, they do not focus on combining the binning with optimizing a regression problem.

Similar to Hyv\"{a}rinen \etal \cite{hyvarinen2016unsupervised}, predicting discrete signal bins has been successfully used for unsupervised video representation learning by discriminating between video segments \cite{dave2022tclr} or classifying video speed \cite{wang2020self}.
Up to the point which the underlying continuous regression function (\eg speed, time, age, depth) can be assumed to be generated by a non-linear combination of non-stationary source (whose statistics change with the function), adding a classification loss decorrelates the independent sources.
Additionally, we hypothesize that there may be a relation between nonlinear ICA and imbalanced sampling: the independent sources $\textbf{s}$ are also continuous and shared across samples. 
And having the hidden representation of the MLP constrained to be independent across dimensions may lead to a better use of sparse samples in certain areas of the target-space.
However, we leave this for future research.

\smallskip \noindent \textbf{Limitations of analysis.}
The analysis performed here is still elementary and only aims to scratch the surface on the usefulness of adding a classification loss when performing regression.
A number of things have been disregarded here such as: 
the effect of the model depth, while keeping the model size fixed.
Additionally, the choice of the optimizer and the loss function during training may also play an important role.
Finally, delving more into the relation between nonlinear ICA and adding a classification loss to the regression, may be an interesting future research avenue.

\section{Related work}
\subsection{Improved deep regression}
A thorough analysis of the effect of deep architecture choices on regression scores is performed in \cite{lathuiliere2019comprehensive}.
Rather than considering architecture choices, other works focus on regression robustness -- defined as less influence from outliers \cite{hastie2015statistical, huber2011robust}.
The benefits of having a weighted regression loss has been extensively analyzed in classic work such as \cite{cleveland1979robust,cleveland1988locally}.
With a similar goal, Barron \cite{barron2019general} proposes a general regression loss that can be adapted to well-known functions.
Minimizing a Tukey\rq s biweight function also offers robustness to outliers \cite{belagiannis2015robust}.
Similarly, a smooth $L_1$ loss is used in \cite{girshick2015fast} for improved object bounding-box regression.
Mapping regression targets to the hyper-sphere can also aid regression \cite{mettes2019hyperspherical}.
Instead of focusing on the loss function, a smooth adaptive activation function is used in \cite{hou2017convnets}.
From a different direction, ensemble networks have been successfully used for improved regression estimates \cite{cortes2014deep, han2016incremental, walach2016learning}.
Deep negative correlation learning is proposed in \cite{zhang2019nonlinear} to learn diversified base networks for deep ensemble regression.
Dissimilar to these works, our goal is not proposing a new network architecture or a new regression loss function, but rather analyzing why a combination between a classification and a regression loss can lead to improvements.

\subsection{Discretized regression targets}
Instead of optimizing continuous object pose angles, \cite{tulsiani2015viewpoints, su2015render} discretize them into classes and perform classification.
Continuous prediction can be obtained back from discretized targets, by using \textsl{soft-min} for disparity learning \cite{kendall2017end}. 
Similarly, ages are discretized in classes and the final prediction is computed as an expectation over class probabilities in \cite{rothe2015dex}. 
Rather than using predefined classes, clusters can be defined and, again, the final predication is a weighted sum over clusters \cite{walker2015dense,wang2015designing}. 
In a similar manner, a weighted combination of clustered regression targets is used for finding surface normals in \cite{wang2015designing}.
Popular object detectors \cite{liu2016ssd,ren2015faster,rogez2017lcr,xiang2017subcategory} also rely on a set of predefined box locations that can be seen as bin centers, and regress the final box locations with respect to these centers.
An explicit combination of regression and classification losses has been shown to improve results for object orientation regression \cite{mousavian20173d,zhao2020monocular,zhou2019objects} by first splitting orientations into bins and then regressing the precise value in each bin.
Similarly, a joint classification loss over discretized targets and a regression loss is effective for horizon line detection \cite{workman2016horizon}.
Here, we want to analyze why these prior works opt for discretizing regression targets. More specifically, we analyze why and when a combination of a regression loss and a classification loss over discretized continuous targets improves results.

\subsection{Regression on imbalanced data}
Prior work has shown that imbalanced sampling can negatively affect the regression scores, and proposed ways to mitigate this by designing better data sampling techniques \cite{branco2017smogn, torgo2013smote}. 
How rare a data point is in the training set, can be modeled with kernel density estimation \cite{steininger2021density}.
Not only the data sampling but also the target distribution can be fixed by smoothing the distribution of both labels and features using nearby targets \cite{yang2021delving}.
Similarly, the learning model can be regularized such that samples that are close in label space are also close in the feature space \cite{gong2022ranksim}.
While focusing on the learning model, ensemble methods are a viable solution when working with imbalanced regression problems \cite{branco2018rebagg}.
Instead of focusing on the data sampling or the model, imbalanced regression estimates can be improved by adapting the MSE (mean squared error) loss \cite{ren2022balanced}.
Most similar to us are \cite{ren2015faster,yang2021delving} whose methods can be seen as defining smooth classes over the data.
However, they aim to improve regression on imbalanced data, while we set off to analyze in which cases adding a classification loss can help regression, and what is the motivation behind this.

\section{Conclusion}

Here, we present a preliminary analysis on the effect of adding a classification loss to the regression loss.
We make the observation that adding a classification loss to the regression loss has been used in computer vision for deep regression \cite{mousavian20173d,workman2016horizon,xiao2019pose}.
And we empirically test across data variations and data samplings on a set of 1D functions, the effect of adding a classification loss to a regression loss.
We find that for imbalanced regression, adding a classification loss helps the most. 
Furthermore, we present an attempt at formalizing this observation starting from the derivations of Ren \etal \cite{ren2022balanced} for imbalanced regression.

Additionally, we validate that adding a classification loss to the regression loss is beneficial on imbalanced real data, where we evaluate on imbalanced image data on NUYD2-DIR and IMDB-WIKI-DIR datasets \cite{yang2021delving}, and imbalanced video progress prediction on the Breakfast dataset \cite{Kuehne12}.  

\small
\smallskip\noindent\textbf{Acknowledgements.} 
This work was done with the support of the Eureka cluster Program, IWISH project, grant number AI2021-066.
Jan van Gemert is financed by the Dutch Research Council (NWO) (project VI.Vidi.192.100).

{\small
\bibliographystyle{ieee_fullname}
\bibliography{egbib}
}
\newpage
\appendix
\onecolumn
\section*{\centering \Large{A step towards understanding why classification helps regression}\\ \normalsize{Supplementary material}}
\section{1D experimental details}
\label{sec:app1d}
\begin{figure}[h!t]
    \begin{tabular}{ccccc}
        {\includegraphics[height=.18\textwidth]{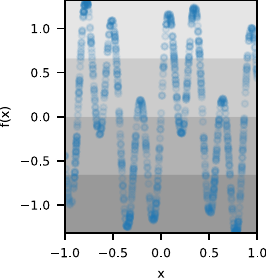}} & 
        {\includegraphics[height=.18\textwidth]{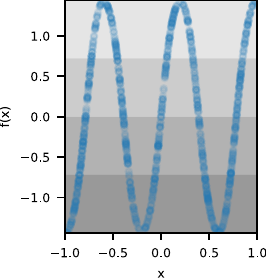}} & 
        {\includegraphics[height=.18\textwidth]{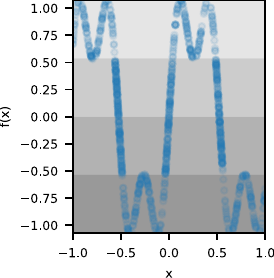}} &
        {\includegraphics[height=.18\textwidth]{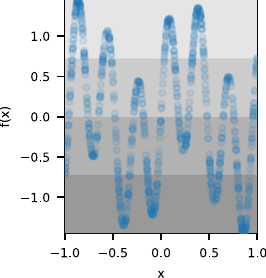}} & 
        {\includegraphics[height=.18\textwidth]{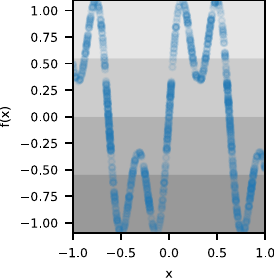}} \\ 
        {\includegraphics[height=.18\textwidth]{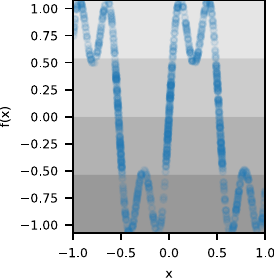}} &
        {\includegraphics[height=.18\textwidth]{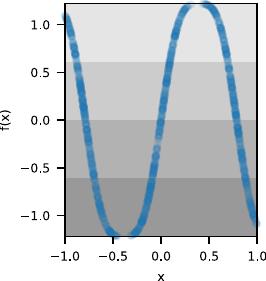}} & 
        {\includegraphics[height=.18\textwidth]{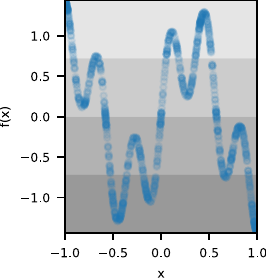}} & 
        {\includegraphics[height=.18\textwidth]{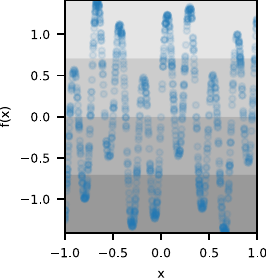}} &
        {\includegraphics[height=.18\textwidth]{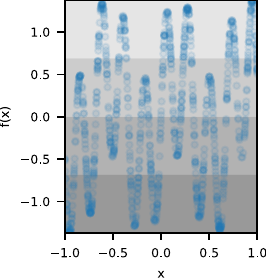}} \\
    \end{tabular}    
    \caption{ \small 
    The 10 sampled 1$D$ functions used in the experiments in the main paper. 
    Over these 10 functions we vary the sampling into: 4 sampling scenarios (\textsl{uniform}, \textsl{mild}, \textsl{moderate} and \textsl{severe}), 3 data scenarios (\textsl{clean}, \textsl{noisy}, and \textsl{ood}), with 5 different seeds and 5 classes ($2^2, 2^4, 2^6, 2^8, 2^{10}$) resulting in 3000 experiments on the 1$D$ data.
    }
    \label{fig:functions}
\end{figure}

For the 1$D$ synthetic experiments we use the function $f(\cdot)$ defined by a sum of two sines with different frequencies and amplitudes:
\begin{align}
 f (x)=a \sin(cx) + b \sin(dx), 
 \label{eq:1d_func}
\end{align}
where $f(x) \in [-1.5, 1.5]$ and $x \in [-1, 1]$.
We show here in \fig{functions} the 10 functions we have sampled from this function for the 1$D$ experiments presented in the paper.

For the validation set we run the experiments over 3 repetitions with seeds: $\{0, 421, 8125\}$, while for testing we run the experiments by setting the random seed everywhere in the code to $\{0, 421, 8125, 2481, 849\}$.
For each dataset scenario (\textsl{clean}, \textsl{noisy data}, \textsl{ood}) and sampling (\textsl{uniform}, \textsl{moderate}, \textsl{mild}, \textsl{severe}) we create 5 dataset versions. 
During training for every random seed we pick a dataset version, giving rise to 5 repetitions per function, over which we report results in the \fig{1d_results} in the paper. 
We print here in \tab{tab1d} the model architecture for the regression-only and regression and classification tasks. 
\begin{table}[t!h]
\centering
\resizebox{0.65\linewidth}{!}{%
    \begin{tabular}{c@{\hskip 0.5in}c}
        \begin{tabular}{ccc}
        \toprule
        Layer  & Output Shape & Param\\ \midrule
        Linear1 &  [-1, 6]     & 12 \\
        ReLU1   &  [-1, 6]     & --- \\              
        Linear2 &  [-1, 16]    & 112 \\
        ReLU2   &  [-1, 16]    & -- \\
        Linear3 &  [-1, 1]     & 17 \\ \bottomrule
        \end{tabular} &
        \begin{tabular}{ccc}
        \toprule
        Layer  & Output Shape & Param\\ \midrule
        Linear1 &  [-1, 6]     & 12 \\
        ReLU1   &  [-1, 6]     & --- \\              
        Linear2 &  [-1, 16]    & 112 \\
        ReLU2   &  [-1, 16]    & -- \\
        Linear3a &  [-1, 1]    & 17\\ 
        \textsl{Linear3b} &  \textsl{[-1, C]}    & \textsl{16C}\\ 
        \bottomrule
        \end{tabular} \\
        (a) Regression only & (b) Regression and classification \\
    \end{tabular} 
    }
    \caption{\small MLP network architectures for the 1D experiments. When adding the classification loss during training, we use an additional linear layer (emphasized here) which maps the features of the one-to-last layer to the desired number of classes, $C$. During testing this layer is discarded.}
\label{tab:tab1d}
\end{table}

\begin{figure}
    \centering
    \centerline{\includegraphics[width=.65\textwidth]{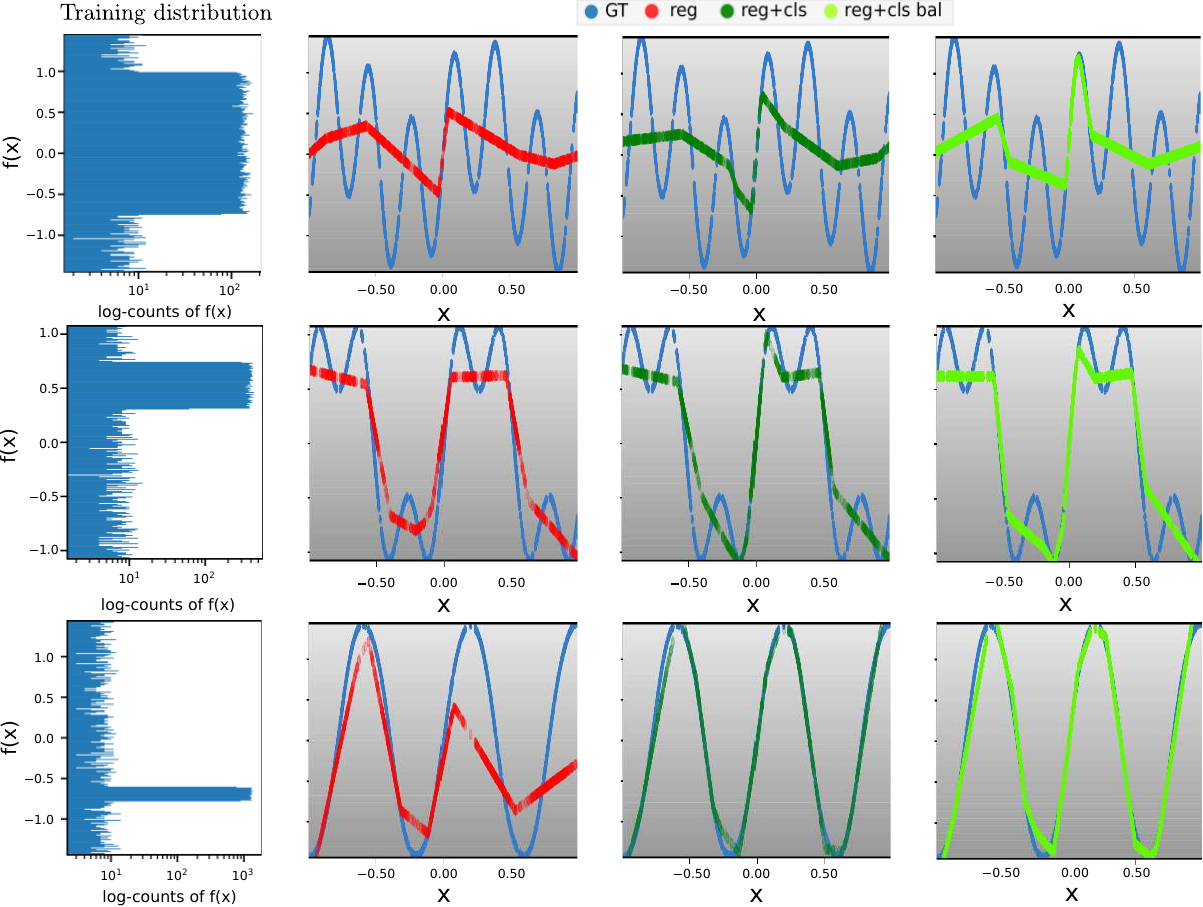}}
    \caption{
       \textbf{Examples of 1$D$ predictions:} 
        For moderate\slash mild\slash severely imbalanced data sampling, on clean data scenario, when using 64 classes.
        The classification variants: with balanced classes (lime) and with imbalanced classes (green) make more accurate predictions than the regression baseline (red).
    }
    \label{fig:bal}
\end{figure}
\fig{bal} shows examples of predictions when using balanced and imbalanced classes on 3 severe sampling scenarios. 
The classification methods can make different mistakes, but overall the predictions are more accurate than for the regression alone.

\begin{figure*}[!ht]
    \centering
    \begin{tabular}{ccc}
        {\includegraphics[width=.3\textwidth]{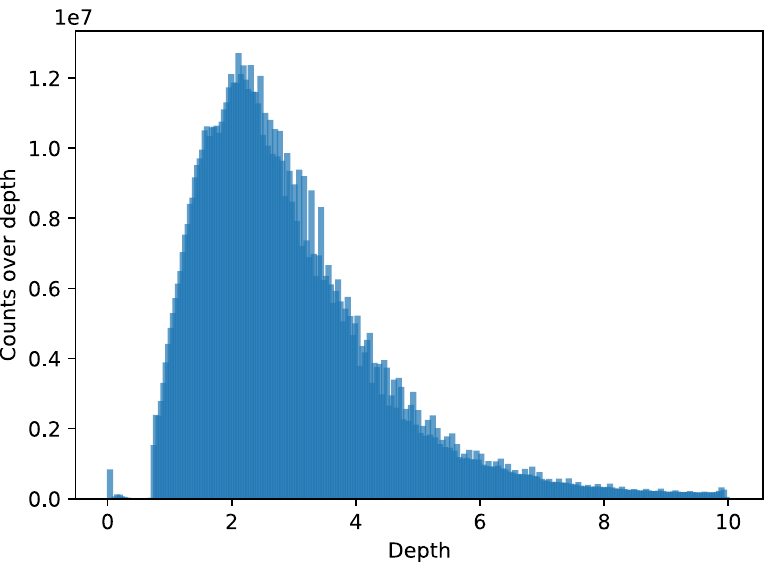}} & 
        {\includegraphics[width=.3\textwidth]{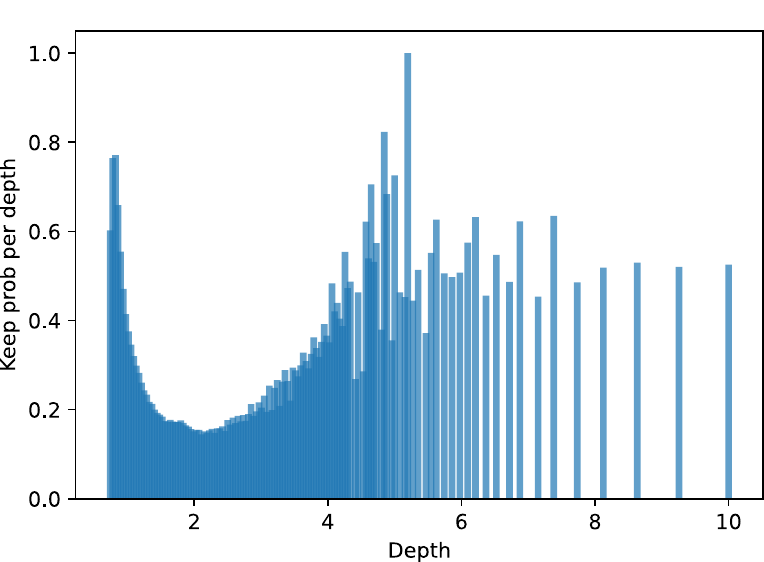}} & 
        {\includegraphics[width=.3\textwidth]{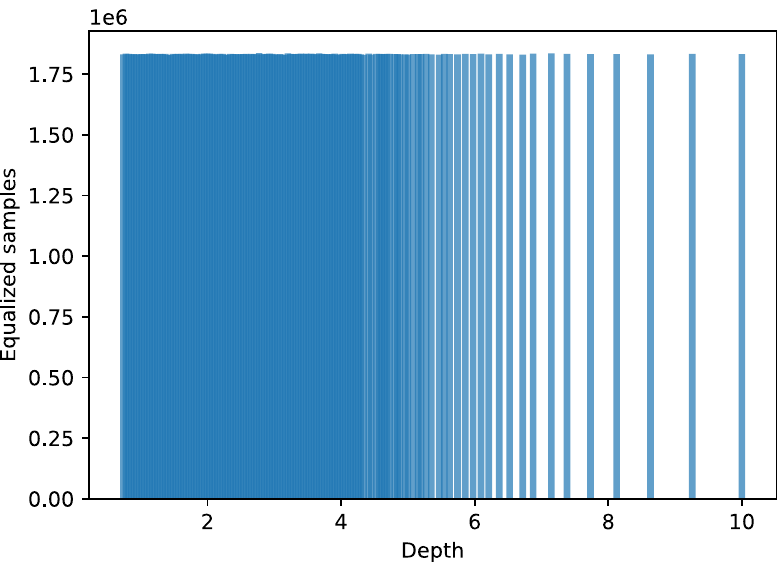}} \\
        \scriptsize{(a) Original NYUD2-DIR targets distribution.} &
        \scriptsize{(b) NYUD2-DIR probabilities of keeping samples per class.} &
        \scriptsize{(c) NYUD2-DIR re-balanced samples.} \\
        {\includegraphics[width=.3\textwidth]{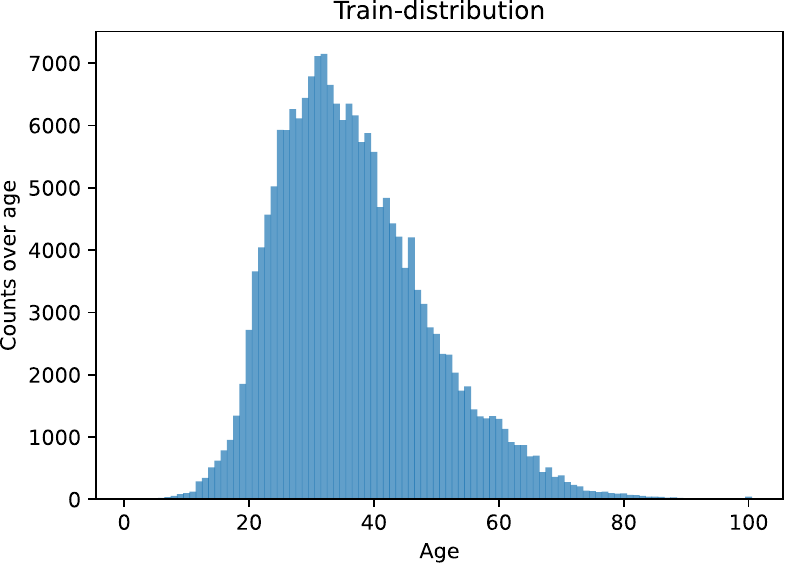}} & 
        {\includegraphics[width=.3\textwidth]{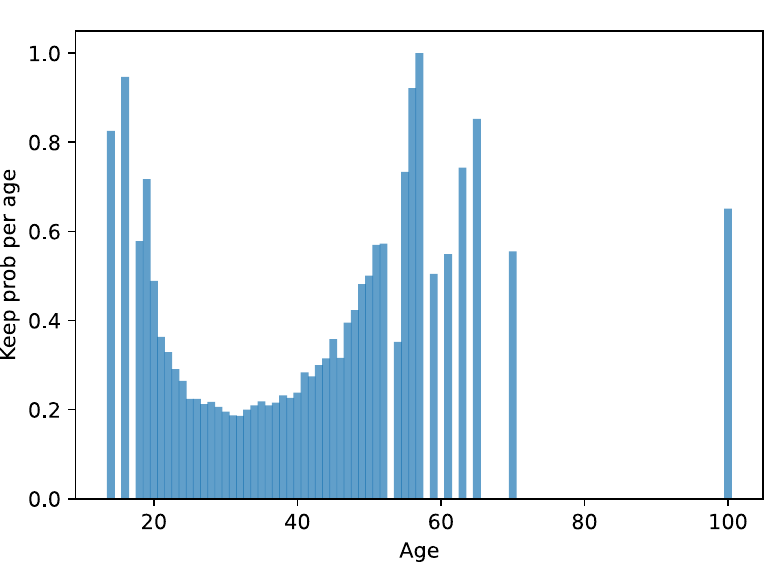}} & 
        {\includegraphics[width=.3\textwidth]{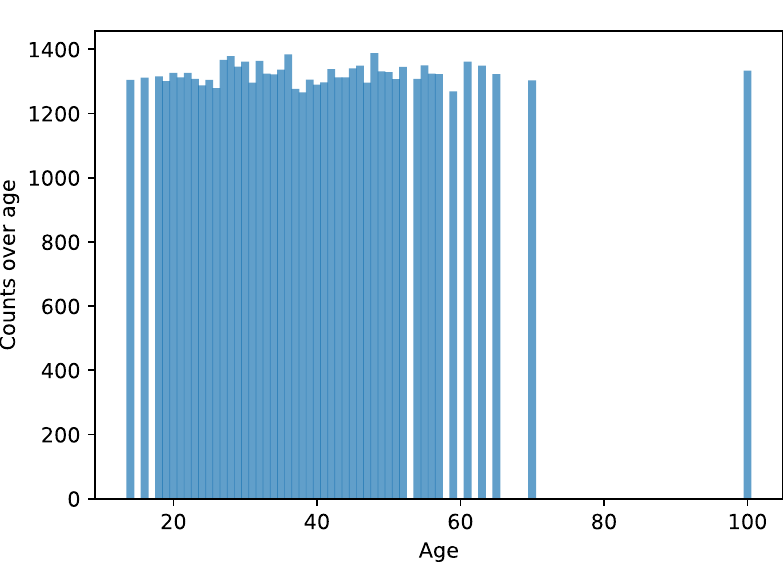}} \\
        \scriptsize{(c) Original IMDB-WIKI-DIR targets distribution.} &
        \scriptsize{(d) IMDB-WIKI-DIR probabilities of keeping samples per class.} & 
        \scriptsize{(e) IMDB-WIKI-DIR re-balanced samples.} \\
    \end{tabular}
    \caption{ \small \textbf{Re-balancing the training distribution in the classification loss.} 
        (a) The initial distribution per class for the NYUD2-DIR dataset.
        (b) The probabilities of samples being retained per class, as described in \eq{bal_keep}.
        (c) The NYUD2-DIR samples re-balanced per class as described in \sect{bal}.
        (d) The initial distribution per class for the IMDB-WIKI-DIR dataset.
        (e) The probabilities of samples being retained per class, as described in \eq{bal_keep}.
        (f) The IMDB-WIKI-DIR samples re-balanced per class as described in \sect{bal}.
        Here we consider 100 classes for binning the dataset targets.
        Equalizing the class ranges using \eq{balance_cls} and sampling samples per class over the training epochs using \eq{bal_keep} is effective at balancing the classes.
        }
    \label{fig:appbal}
\end{figure*}
\section{Re-balancing the 2D imbalanced training data}
\label{app:bal_ex}
For re-balancing the classes, we first redefine the class ranges to have equalized class counts using \eq{balance_cls}. 
For the regression loss we use all samples in each batch during training.
For the classification loss we subsample the samples in the batch by considering the current training sample $(\mathbf{x}, y_k)$ in the classification loss with probability $\rho(k)$, where $\rho(k)$ is defined in \eq{bal_keep}.
Specifically, for every training sample $(\mathbf{x}, y_k)$ we sample a random variable $u$ from the uniform distribution and use this sample in the classification loss if the random value is smaller than $\rho(k)$:
\begin{align}
    & u \sim \mathcal{U}(0,1),\\
    \text{we classify }(\mathbf{x}, y_k) \text{ if } & u \le \rho(k).
\end{align}
This procedure is useful, because the complete dataset will be visited during training for classification (provided sufficient epochs), yet the samples seen per class will be equal.
To illustrate this, we show in \fig{appbal}(a) the original data distribution on NYUD2-DIR, and IMDB-WIKI-DIR, the probabilities per class of keeping a sample for 100 equalized classes in \fig{appbal}(b), and the re-balanced training set distribution in \fig{appbal}(c). 
Here we consider 100 classes to bin the training targets.
We see that after class equalization and re-balancing the class distribution is uniform. 
We could have only used the second option to balance the classes by sampling uniform samples per class using \eq{bal_keep} without first equalizing the classes with \eq{balance_cls}. 
However, if the classes are severely imbalanced (as in the case of our data), some class probabilities are extremely low which leads to never selecting samples from those classes, and thus having the classification fail to converge.

\section{Breakfast dataset task variations}
\label{app:bf_ex}
\begin{figure*}[!ht]
    \centering
\resizebox{1.\linewidth}{!}{%
    \begin{tabular}{ccccc}
        {\includegraphics[width=.2\textwidth]{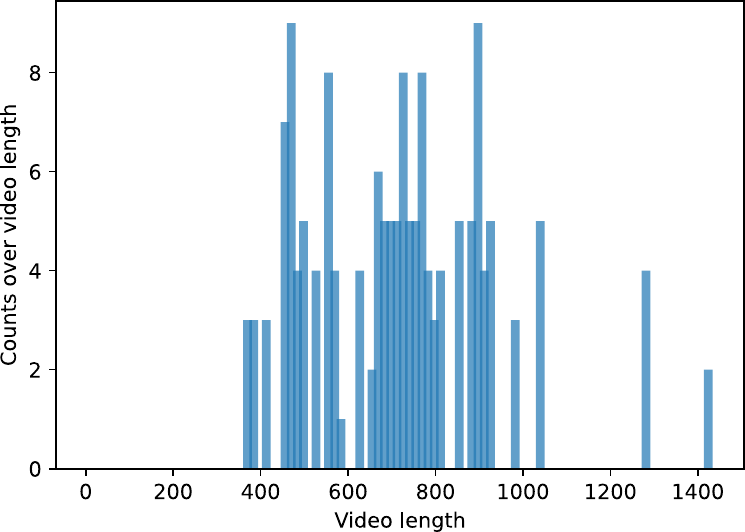}} & 
        {\includegraphics[width=.2\textwidth]{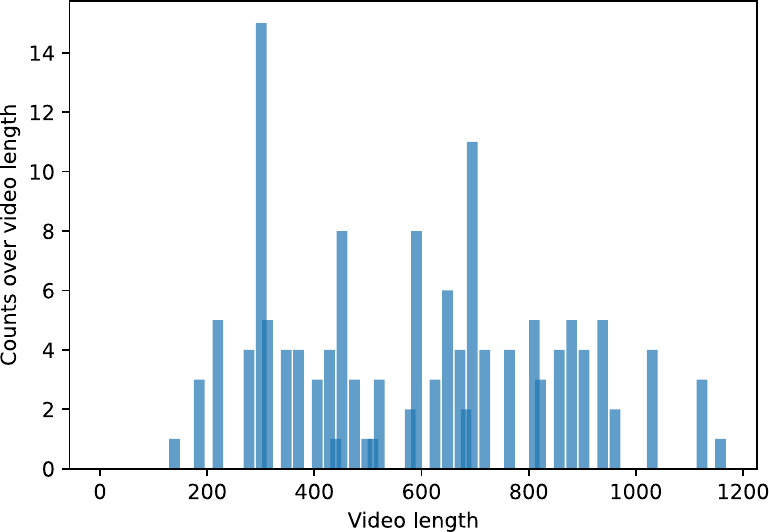}} &
        {\includegraphics[width=.2\textwidth]{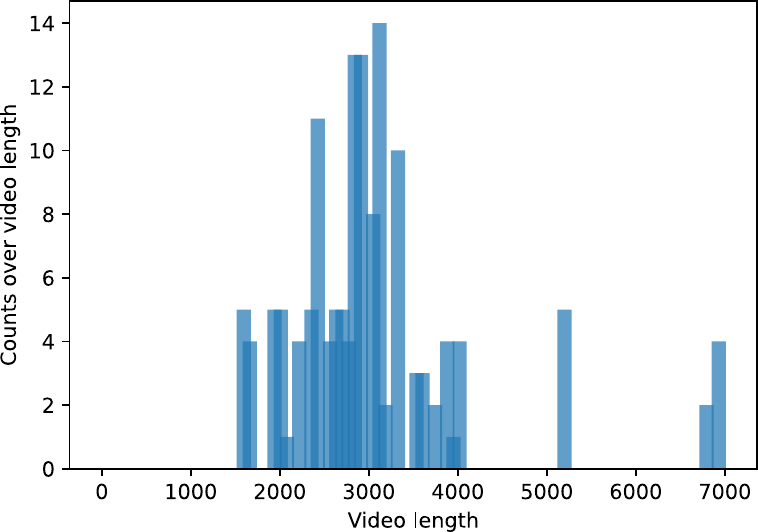}} &
        {\includegraphics[width=.2\textwidth]{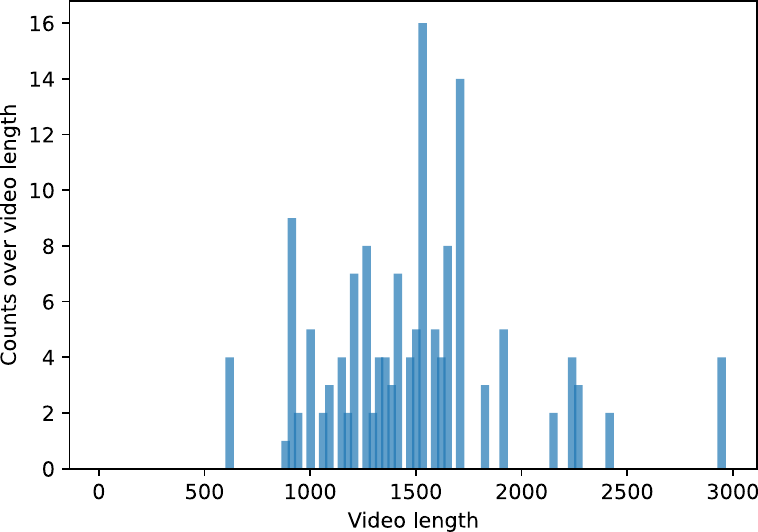}} & 
        {\includegraphics[width=.2\textwidth]{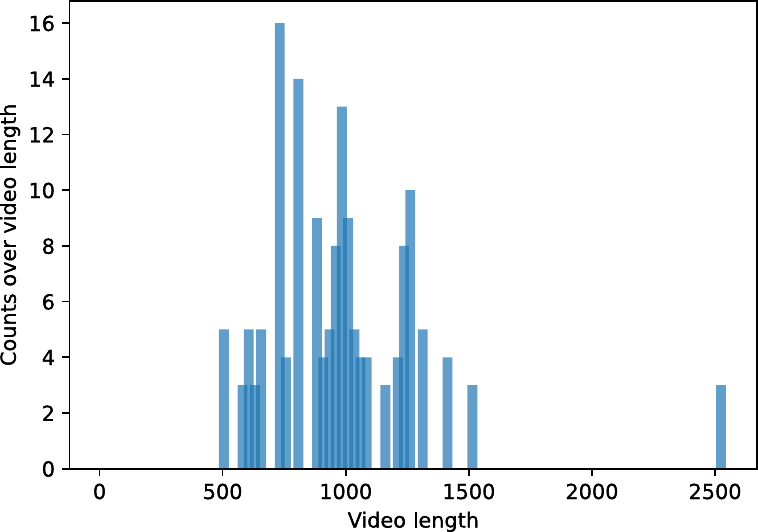}} \\ 
        \scriptsize{(a) Cereals} &
        \scriptsize{(b) Coffee} &
        \scriptsize{(c) Fried egg} &
        \scriptsize{(d) Juice} &
        \scriptsize{(e) Milk} \\
        {\includegraphics[width=.2\textwidth]{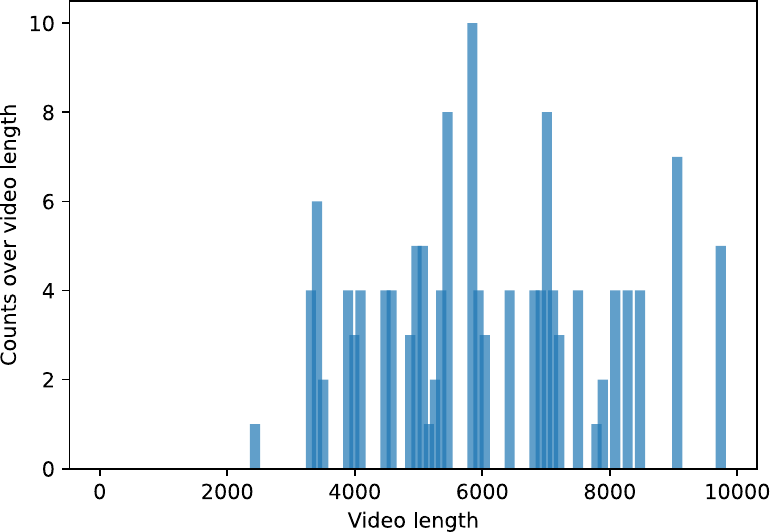}} &
        {\includegraphics[width=.2\textwidth]{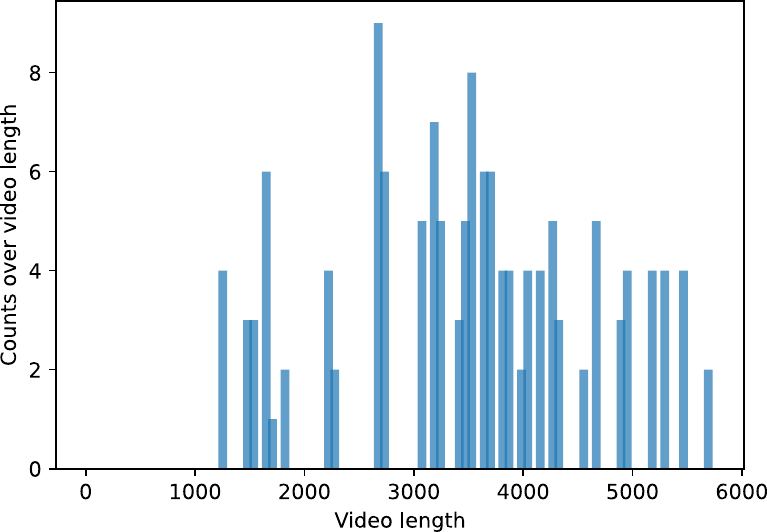}} & 
        {\includegraphics[width=.2\textwidth]{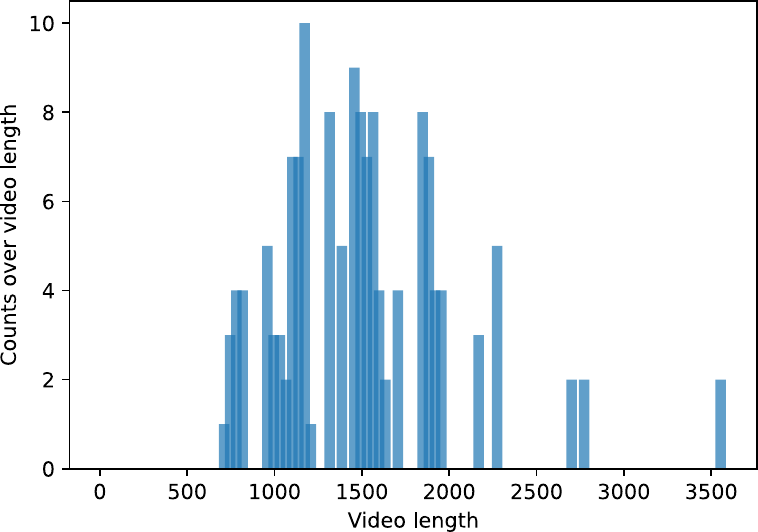}} & 
        {\includegraphics[width=.2\textwidth]{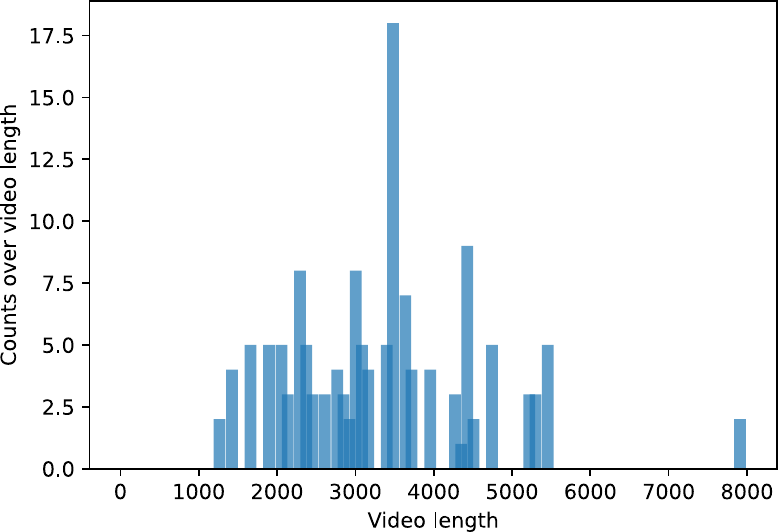}} & 
        {\includegraphics[width=.2\textwidth]{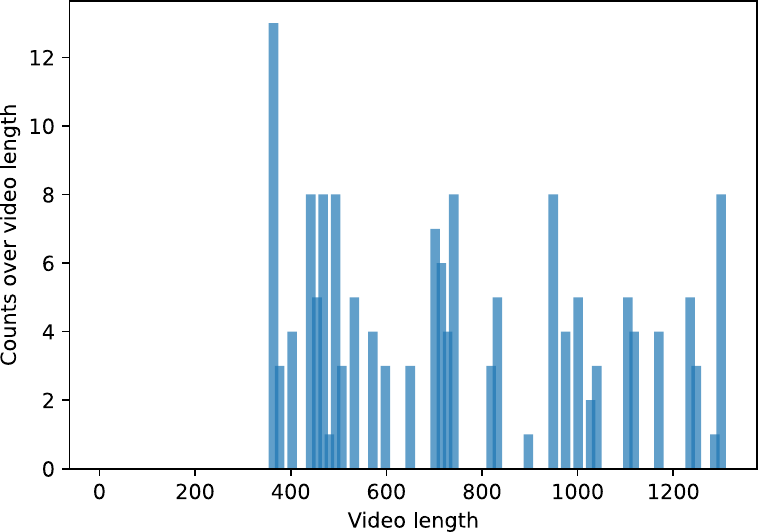}}\\
        \scriptsize{(a) Pancake} &
        \scriptsize{(b) Salad} &
        \scriptsize{(b) Sandwich} &
        \scriptsize{(c) Scrambled egg} &
        \scriptsize{(d) Tea} \\
    \end{tabular}
}
    \caption{ \small \textbf{Breakfast task length variations.} 
        Here we consider 100 classes for binning the video lengths.
        There is a considerable variation in the video lengths for the same cooking task. 
        Because of this imbalance in the training data, the model makes more errors when predicting video progression on videos that are either considerably longer or shorter than the average.
        Next to this, the large variety in appearance of the videos makes the video progress prediction challenging.
    }
    \label{fig:appbal2}
\end{figure*}

\fig{appbal2} shows the large variation in video lengths across the 10 tasks on the Breakfast dataset on the training data. 
This large variability makes predicting video progression extra challenging, even when predicting video progression in percentages.
This is due to having a most-frequent task length, and the model learns to predict better for the videos belonging to the most-frequent task length, and it makes mistakes when predicting progression on the videos that are a lot shorter or a lot longer than the average. 



\end{document}